\pdfoutput=1
\PassOptionsToPackage{table,dvipsnames}{xcolor}

\documentclass[11pt]{article}
\usepackage[]{ACL2023}

\usepackage{multirow}
\usepackage{amsmath}
\usepackage{amssymb}
\usepackage{times}
\usepackage{latexsym}
\usepackage[T1]{fontenc}
\usepackage[utf8]{inputenc}
\usepackage{microtype}
\usepackage{inconsolata}

\usepackage{tcolorbox}
\usepackage{enumitem}

\usepackage{graphicx} 
\usepackage{multirow}
\usepackage{threeparttable}
\usepackage{booktabs}

\usepackage{fontawesome5}
\definecolor{Gold}{RGB}{255,215,0}
\definecolor{Silver}{RGB}{192,192,192}
\definecolor{Bronze}{RGB}{205,127,50}

\title{Neural at ArchEHR-QA 2025: Agentic Prompt Optimization for Evidence-Grounded Clinical Question Answering}

\author{%
  Sai Prasanna Teja Reddy\textsuperscript{1}, 
  Abrar Majeedi, 
  Viswanatha Reddy Gajjala,\\
  \textbf{Zhuoyan Xu}, 
  \textbf{Siddhant Rai}, 
  \textbf{Vaishnav Potlapalli}\\[1ex]
  \textsuperscript{1}University of Chicago\\
  \texttt{bogireddyteja@uchicago.edu}
}

\definecolor{verylightgray}{gray}{0.9}

\begin{document}
\maketitle

\begin{abstract}

Automated question answering (QA) over electronic health records (EHRs) can bridge critical information gaps for clinicians and patients, yet it demands both precise evidence retrieval and faithful answer generation under limited supervision. In this work, we present \emph{Neural}, the runner‑up  in the BioNLP 2025 ArchEHR‑QA shared task on evidence‑grounded clinical QA. Our proposed method decouples the task into (1) sentence‑level evidence identification and (2) answer synthesis with explicit citations. For each stage, we automatically explore the prompt space with DSPy’s MIPROv2 optimizer, jointly tuning instructions and few‑shot demonstrations on the development set. A self‑consistency voting scheme further improves evidence recall without sacrificing precision. On the hidden test set, our method attains an overall score of 51.5, placing second stage while outperforming standard zero‑shot and few‑shot prompting by over 20 and 10 points, respectively. These results indicate that data‑driven prompt optimization is a cost‑effective alternative to model fine‑tuning for high‑stakes clinical QA, advancing the reliability of AI assistants in healthcare.


\end{abstract}

\section{Introduction}

Automatically generating answers to patients’ medical questions using information from their electronic health records (EHRs) poses significant challenges, but also offers substantial potential for improving clinical communication and patient engagement \cite{soni-etal-2025-archehr-qa}. The ArchEHR-QA 2025 shared task directly targets this problem by providing patient questions alongside excerpts from clinicians’ notes, and requiring systems to generate grounded responses that explicitly cite the supporting sentences.

Recent advances in Large Language Models (LLMs) have shown promising results in the domain of answering clinical questions based on unstructured patient notes \cite{singhal2025toward}. However, fine-tuning LLMs for answering clinical questions based on unstructured patient notes is constrained by the limited availability of supervised clinical data, which increases the risk of overfitting. Consequently, prompt-based methods offer a practical and cost effective solution. Despite the advantages, prompt engineering comes with its own set of challenges \cite{karayanni2024keeping}. Crafting effective prompts for complex tasks often requires expert effort and iterative refinement. This difficulty is amplified in the clinical domain, where identifying the correct evidence from lengthy medical notes is critical for accurate answers. Prior studies have explored manual prompt designs and chain-of-thought cues for medical QA \cite{tai2025clinical}, yet these ad-hoc methods may not yield optimal performance. Automated prompt optimization techniques \cite{wang2023promptagent} offer a systematic alternative, but often treat each task holistically and may not incorporate domain expertise effectively.

In this work, we introduce a two-stage LLM pipeline for clinical question answering that explicitly separates evidence identification and answer generation. In each stage, prompts are automatically optimized using the MIPROv2 optimizer from DSPy ~\cite{khattab-etal-2021-relevance,khattab2024dspy}. The first stage is dedicated to identifying the relevant information within the clinical note, while the second stage leverages this information to generate a precise and contextually appropriate answer. By decomposing the task into these two distinct stages, it becomes possible to define clear, stage-specific evaluation objectives, namely, F1 score for evidence retrieval performance and the mean of word limit score, citation format score, BLEU, ROUGE, SARI, BERT, ALIGN, and MEDCON scores for answer quality metrics. This decomposition also enables the use of optimization algorithms to systematically search for prompts that maximize these objectives. To further improve reliability, we integrate a self-consistency \cite{wang2022self} approach in the evidence retrieval stage: the model is run multiple times, and a majority vote over the outputs determines the final cited sentences, reducing variability and errors.

In summary, our contributions are: 

\begin{itemize}
  \item \textbf{Decomposed Prompt Optimization Framework:} We propose a two-stage pipeline that modularizes clinical QA, enabling distinct and targeted prompt optimization for evidence retrieval and answer generation, a paradigm shift from monolithic optimization approaches.
  
  \item \textbf{Systematic Instruction Space Exploration:} We demonstrate the efficacy of leveraging advanced optimizers like MIPROv2 to discover high-performing, task-specific prompt configurations from limited development data, enhancing both performance and reproducibility.

  \item We perform a rigorous evaluation on an expert‑annotated clinical QA dataset, demonstrating that our prompt‑optimized pipeline yields significant improvements in factual accuracy and answer relevance compared to established baselines, underscoring its effectiveness for reliable clinical-QA.
  
\end{itemize}

\section{Related Work}

\paragraph{Clinical QA:} Developing QA systems for clinical data has long been an interest in biomedical NLP. Earlier datasets like emrQA \cite{pampari2018emrqa} generated large-scale QA pairs from electronic medical records by repurposing annotations, but these often contained synthetic questions or required mapping to structured outputs. Recent research has shown that large LLMs can achieve near-expert performance on medical QA benchmarks \cite{singhal2025toward}.

\paragraph{Prompt Optimization:} There is a growing interest in automated prompt search or optimization. More recently, methods such as APE \cite{zhou2022large} and OPRO \cite{yang2023large} treat prompt design as a black-box optimization problem, iteratively refining prompts by evaluating model outputs. MIPRO \cite{opsahl2024optimizing} extends this idea to multi-stage LLM programs, jointly optimizing the instructions and demonstration examples of each module in a pipeline. Our work leverages the latest optimizer, MIPROv2 \cite{opsahl2024optimizing}, which uses a combination of prompt proposal and Bayesian search to find high-performing prompts efficiently.

\paragraph{Self-Consistency:} Large LLMs can produce variable outputs given the same prompt, especially under chain-of-thought reasoning. The self-consistency decoding strategy \cite{wang2022self} addresses this by sampling multiple outputs and choosing the result that is most consistent across samples.

\section{Methodology}

Our method draws on a human‐inspired decoupling strategy, separating evidence gathering from solution formulation. In Stage 1, we identify relevant resources analogous to conducting a web search or literature review by retrieving key sentences. In Stage 2, we frame the final solution by synthesizing insights from the retrieved facts. We operationalize this intuition as a modular, two‐stage pipeline tailored to clinical QA.

Consider each clinical note excerpt is segmented into individual sentences $s_1, s_2, \dots, s_n$, and each sentence $s_i$ is annotated with a label $y_{i}$ $\in  (\text{essential}, \text{not-relevant}, \text{supplementary})$. The label indicates whether $s_i$ contains information essential for answering a given patient/clinician question $q$. This sentence-level annotation forms the basis of Stage-1, while Stage-2 uses the content of the essential sentences (post consistency testing) to produce the final answer $a_{gen}$.

\subsection{Sentence-Level Essentiality Classification}


For a question–note pair let
\[
Y^{+} \;=\; \bigl\{\,i \,\bigm|\, y_i = 1 \bigr\},
\text{and} \\~~
\hat{Y}^{+} \;=\; \bigl\{\,i \,\bigm|\, \hat{y}_i = 1 \bigr\},
\]
denote, respectively, the indices of \emph{gold‑standard essential} sentences and the indices predicted essential by the model. We begin with a manually crafted prompt that presents the question~\(q\) and the sentence sequence \(\{s_1,\dots,s_n\}\) and requests a binary relevance label for every sentence in addressing the \(q\).

\paragraph{Prompt‑Optimization Objective (Stage-1):} We invoke the \textbf{MIPROv2} to optimize the prompt. Treating the instruction text (and any embedded demonstrations) as discrete parameters \(P\in\mathcal{P}\), MIPROv2 iteratively: (i) proposes a candidate prompt~\(P\), (ii) applies the fixed LLM to the training set, and (iii) updates \(P\) so as to \emph{maximize} the sentence‑level \(F_{1}\!\bigl(Y^{+},\hat{Y}^{+}\bigr)\). By searching this space of instructions and few‑shot exemplars, the optimizer converges on a prompt \(P^{*}\) that elicits labels with markedly higher precision and recall, thereby yielding a more reliable evidence set for Stage~2.




\paragraph{Self‑Consistency Voting:}
To improve the reliability of Stage~1, we apply a \emph{self‑consistency voting} scheme: the classifier is executed \(R=5\) times on the same
\((q,\{s_i\})\) input, each run differing only in its stochastic seed. Let \(\hat{y}_{i}^{(r)}\in\{0,1\}\) be the binary prediction for sentence \(s_i\) in run \(r\) (\(1=\text{essential}\)). The final label is obtained by majority vote,

\[
v_{i} \;=\; \sum_{r=1}^{R}\hat{y}_{i}^{(r)},
\quad
\hat{y}_{i} \;=\;
\begin{cases}
1 & \text{if } v_{i} \ge \tau=\lceil R/2\rceil,\\[4pt] 
0 & \text{otherwise},
\end{cases}
\quad
\]

This aggregation suppresses spurious single‑run errors and retains sentences identified as essential by at least three of the five passes, thereby reducing variance and boosting the expected \(F_{1}\) of the evidence selection step.

\subsection{Answer Generation from Essential Sentences}


Let \(q\) be the input question and let
$
E \;=\; \bigl\{\, s_i \;\bigm|\; \hat{y}_i = 1 \bigr\}
$
denote the set of sentences that Stage~1 predicted as \emph{essential}. Given the pair \((q,E)\), Stage~2 must produce a concise natural-language answer \(a_{\text{gen}}\) that (i) directly addresses \(q\), (ii) contains at most \(75\) words, and (iii) cites the supporting sentences in \(E\) using the required parenthetical notation. We initialise Stage~2 with a hand‑written prompt template and then invoke
\textbf{MIPROv2} to optimize this template.
Let \(P\) denote a prompt parameterised by its instruction wording and any embedded demonstrations, and let \(g_{\theta}(\,\cdot\,;P)\) be the fixed LLM generator. Given an input pair \((q,E)\) the model outputs $
a_{\text{gen}}
\;=\;
g_{\theta}\!\bigl( (q,E); P \bigr)$.

\paragraph{Prompt‑Optimization Objective (Stage-2):} The goal is to maximise the composite reward
\[
\begin{gathered}
\mathcal{R}\bigl(a_{gen}, a^{*}, E\bigr)
= \underbrace{\mathbf{1}\!\bigl[\lvert a_{gen}\rvert \le 75\bigr]}_{\text{length}}
\\[4pt]
+\,\underbrace{\mathbf{1}\!\bigl[\text{format}(a_{gen},E)\bigr]}_{\text{citations}}
\;+\;
\underbrace{\tfrac16\displaystyle\sum_{m\in\mathcal{M}}
           m\!\bigl(a_{\text{gen}}, a^{*}\bigr)}_{\text{surface \& semantic quality}}
\end{gathered}
\]
where \(a^{*}\) is the reference answer,
\(\lvert\,\cdot\,\rvert\) counts words,
and
\[
\begin{gathered}
\mathcal{M} \;=\;
\bigl\{\,\text{BLEU},\; \text{ROUGE},\; \text{SARI},\\
       \text{BERT},\; \text{Align},\; \text{MEDCON}\bigr\}.
\end{gathered}
\]

The indicator terms enforce hard constraints on length and citation format, while the mean of the six metrics rewards lexical overlap, semantic fidelity, factual consistency, and medical--concept coverage.

\paragraph{Search Procedure:} MIPROv2 explores the discrete prompt space \(\mathcal{P}\) by iteratively proposing candidate prompts, evaluating them on a validation set, and selecting
\[
P^{*}= \arg\!\max_{P \in \mathcal{P}}
\;
\mathbb{E}_{(q,E,a^{*})}
\Bigl[
\mathcal{R}\!
\bigl(
  g_{\theta}((q,E);P),
  a^{*},
  E
\bigr)
\Bigr]
\]
The optimal prompt \(P^{*}\) consistently elicits answers that are succinct,
properly cited, and of high quality according to all surface--level and
semantic metrics, thus satisfying the Stage~2 requirements.

\begin{table*}[!t]
  \caption{
    Evaluation of participants on factuality and relevance metrics. 
    \textbf{Bold} indicates the best performance in each column, \underline{underlined} the second best. 
    Here $P^{S}$, $R^{S}$, $F_{1}^{S}$ denote micro‑averaged \emph{strict} precision, recall and F1; 
    $P^{L}$, $R^{L}$, $F_{1}^{L}$ denote micro‑averaged \emph{lenient} precision, recall and F1; 
    AVG$_\mathrm{fact}$ and AVG$_\mathrm{relev}$ are the official “Overall Factuality” and “Overall Relevance” scores, and “Overall” is the combined score. Abbreviations: R.L.=
ROUGE-Lsum, B.S. = BERTScore, A.S. = AlignScore, M.C. = MEDCON . }
  \label{tab:fact_relev}
  \centering
  \begin{threeparttable}
    \begin{small}
      \renewcommand{\multirowsetup}{\centering}
      \setlength{\tabcolsep}{4.2pt}
      \begin{tabular}{l|*{7}{c} | *{7}{c}|c}
        \toprule
        \multirow{2}{*}{Model}
          & \multicolumn{7}{c|}{\textbf{Factuality}}
          & \multicolumn{7}{c|}{\textbf{Relevance}}
          & \multirow{2}{*}{Overall} \\
        \cmidrule(lr){2-8} \cmidrule(lr){9-15}
  & \scalebox{0.78}{$P^{S}$}
  & \scalebox{0.78}{$R^{S}$}
  & \scalebox{0.78}{$F_{1}^{S}$}
  & \scalebox{0.78}{$P^{L}$}
  & \scalebox{0.78}{$R^{L}$}
  & \scalebox{0.78}{$F_{1}^{L}$}
  & \scalebox{0.78}{AVG$_\mathrm{fact}$}
  & \scalebox{0.78}{BLEU}
  & \scalebox{0.78}{R.L.}
  & \scalebox{0.78}{SARI}
  & \scalebox{0.78}{B.S.}
  & \scalebox{0.78}{A.S.}
  & \scalebox{0.78}{M.C.}
  & \scalebox{0.78}{AVG$_\mathrm{relev}$} \\
\midrule

        DMISLab {\textcolor{Gold}{\faTrophy}}  
          & 57.9   & 59.3   & 58.6   
          & 61.2   & 59.2   & 60.2   
          & 58.6   
          & \textbf{14.3}   & \textbf{46.5}   & 36.7   & \textbf{53.9}   & \textbf{92.4}   & \textbf{49.3}   & \textbf{48.8}   
          & \textbf{53.7}   \\

        \cellcolor{verylightgray} \textbf{Ours} {\textcolor{Silver}{\faTrophy}} 
          & \cellcolor{verylightgray}55.4
          & \cellcolor{verylightgray}\underline{63.8}
          & \cellcolor{verylightgray}\underline{59.3}
          & \cellcolor{verylightgray}58.4
          & \cellcolor{verylightgray}\underline{63.7}
          & \cellcolor{verylightgray}\underline{60.9}
          & \cellcolor{verylightgray}\underline{59.3}
          & \cellcolor{verylightgray}\underline{8.5}
          & \cellcolor{verylightgray}\underline{34.1}
          & \cellcolor{verylightgray}\textbf{73.1}
          & \cellcolor{verylightgray}\underline{39.1}
          & \cellcolor{verylightgray}67.3
          & \cellcolor{verylightgray}40.0
          & \cellcolor{verylightgray}\underline{43.7}
          & \cellcolor{verylightgray}\underline{51.5} \\
  
        LAILab  {\textcolor{Bronze}{\faTrophy}} 
          & 56.0   & \textbf{65.5}     & \textbf{60.4}   
          & 59.7   & \textbf{66.0}     & \textbf{62.7}   
          & \textbf{60.4}   
          & 6.5    & 32.7    & 69.2    & 37.4    & 65.3    & 38.4    & 41.6   
          & 51.0            \\

        LAMAR 
          & \underline{60.6}   & 53.6   & 56.9   
          & \underline{64.0}   & 53.5   & 58.3   
          & 56.9   
          & 6.0    & 32.1    & 65.8    & 36.4    & 64.3    & \underline{43.6}    & 41.4   
          & 49.1            \\

        ssagarwal 
          & 68.8   & 36.2   & 47.5   
          & 71.7   & 35.6   & 47.6   
          & 47.5   
          & 4.7    & 31.1    & \underline{70.0}   & 36.9    & \underline{74.9}   & 38.0    & 42.6   
          & 45.0            \\
          \midrule
        Few-Shot 
          &  71.2  &  38.2  &  49.8
          &  74.5  & 37.8   &   50.2
          &  49.8  
          &  1.7   &  25.5   &  53.9  &  28.7  &  54.5  &  39.7   &   34.0
          &  41.9        \\

        Zero-Shot 
          & \textbf{71.6}  & 21.9   & 33.6   
          & \textbf{77.0}   & 22.3   & 34.6  
          & 33.6  
          & 0.1    & 15.2    & 47.8   & 20.5    & 57.7   & 25.6    & 27.8  
          & 30.7            \\
        \bottomrule
      \end{tabular}
    \end{small}
  \end{threeparttable}
  \vspace{-1.5em}
\end{table*}

\section{Experimental Setup}

\paragraph{Dataset:} We evaluated our system on the ArchEHR-QA 2025 dataset \cite{soni-etal-2025-dataset-patient-needs}. This dataset contains 120 question-note cases derived from MIMIC-III/IV clinical notes. Each case includes a patient question (often a layperson’s phrasing) and a clinician-rewritten question focusing on the key medical query, along with a relevant excerpt from the patient’s EHR notes. The notes are annotated with sentence numbers and labels indicating relevance (“essential,” “supplementary,” “not relevant”) to the question. The official split provides 20 cases as a development set and 100 cases as a test set . We used the 20 development cases for prompt optimization and for all ablations. Final results on the test set were obtained via the Codabench submission system. 

\paragraph{Evaluation Metrics:}

Following the official ArchEHRQA shared task protocol, we evaluate each submission along two complementary axes: \emph{Factuality} and \emph{Relevance}, which help capture evidence faithfulness and response quality. \emph{Factuality} is quantified by matching the set of note sentences cited by the model against expert‑annotated evidence and computing precision, recall, and F1. We report a \textbf{strict} variant that counts only \emph{essential} citations and a \textbf{lenient} variant that also accepts \emph{supplementary} evidence, following the task guidelines. \emph{Relevance} is evaluated as the arithmetic mean of complementary surface and semantic level metrics: BLEU ~\cite{papineni2002bleu}, ROGUE ~\cite{lin-2004-rouge}, SARI ~\cite{xu2016optimizing}, BERTScore ~\cite{zhang2020bertscoreevaluatingtextgeneration}, AlignScore \cite{zha2023alignscore}, and MEDCON \cite{yim2023aci}.







\paragraph{Baselines:} To gauge the performance of our prompt optimization approach, we compare it against two baselines:

\begin{itemize}

\item \textbf{Zero-Shot Prompting:} A single, succinct instruction per stage. This reflects the common practice of “plug‑and‑play” prompting without any exemplars.

\item \textbf{Few-Shot Prompting:} Adds two manually selected demonstrations to each stage’s prompt but preserves the terse directive style. This isolates the value of exemplars alone, without optimization.


\end{itemize}

\section{Results}

Table~\ref{tab:fact_relev} presents the comparative performance of our system alongside competing submissions on the ArchEHR-QA 2025 test set. Our approach ranked \textbf{second overall}, achieving a combined score of \textbf{51.5}, with individual scores of \textbf{59.3} for factuality and \textbf{43.7} for relevance. Crucially, our system maintained \textbf{consistently high performance across all evaluation axes}, in contrast to other systems that exhibited strong performance on isolated metrics but lacked robustness overall. We observe a \textbf{substantial margin of improvement over baseline prompting strategies}: our method outperforms the \textbf{zero-shot} and \textbf{few-shot} variants by approximately \textbf{20 and 10 points}, respectively, on the overall score. These gains underscore the effectiveness of \emph{automated prompt optimization}, which systematically discovers high-performing instructions and demonstrations tailored to each stage of the QA pipeline. Moreover, our system’s relative stability across metrics—including both surface-level (BLEU, ROUGE, SARI) and semantic (BERTScore, AlignScore, MEDCON) relevance measures—suggests that \textbf{prompt optimization not only improves individual metrics but also contributes to the holistic quality and trustworthiness of generated answers}. These findings affirm our central claim: that prompt optimization is not merely a heuristic tuning step, but a principled and impactful method for enhancing LLM-based clinical QA systems.

\section{Conclusion}

We propose a two-stage approach for clinical question answering on medical notes, leveraging DSPy’s MIPROv2 optimizer to autonomously fine-tune prompts for each stage. In Stage 1, the method extracts essential evidence from the notes by optimizing the prompt to maximize the evidence F1 score. 
In Stage 2, the system generates answers by optimizing a prompt based on a composite metric incorporating several metric (word limit score, citation format score, BLEU, ROUGE, etc.),
yielding concise, structured, and clinically reliable response. This prompt-optimized pipeline demonstrates substantial improvements over baselines, highlighting the efficacy of prompt optimization within a modular LLM framework. The results suggest that prompt engineering can transit from heuristic practice to data-driven optimization process, identifying high-performing prompts tailored to specific tasks. For medical question answering systems, this advancement enhances both evidence retrieval and answer trustworthiness, representing a significant step toward the development of reliable AI assistants for clinicians and patients.

Future research directions include integrating web search agents to retrieve external medical knowledge absent from clinical notes, further enriching the capabilities and completeness of automated clinical QA systems. 

\section{Limitations}

Despite strong performance on the ArchEHR-QA benchmark, our two-stage prompt-optimized framework faces limitations rooted in both data and model design. The curated and annotated EHR excerpts used for evaluation do not reflect the messiness of real-world clinical notes, which often suffer from incompleteness, inconsistency, and institutional variability; this makes generalization across healthcare settings difficult, especially given the lack of standardization and privacy restrictions on accessing realistic data. Furthermore, the model has not been domain-adapted and relies on a generic tokenizer, potentially missing specialized medical vocabulary crucial for understanding nuanced queries. The modular two-step process, while flexible, introduces latency and risk of compounding errors, especially as the size of the candidate space in MIPROv2 grows. This reranker also depends heavily on metrics like BLEU, which can reward surface-level similarity over true semantic alignment and are sensitive to the distribution of training data. Together, these factors raise concerns about both scalability and the quality of alignment, even when evaluation scores appear strong.

\section{LLM Settings}

In both stages of our pipeline—sentence‐level evidence identification and answer synthesis—we employ the GPT-4.1 model accessed via the OpenAI API. To accommodate the extensive clinical context and few‐shot demonstrations during prompt optimization, we allocate a maximum context window of 10,000 tokens. All prompt‐optimization experiments (i.e., MIPROv2’s evaluation of candidate prompt templates and few‐shot exemplars) are conducted with a low‐variance decoding strategy, setting the temperature to 0.3. This relatively “cold” sampling regime promotes determinism, ensuring that our optimizer receives consistent feedback on prompt efficacy as measured by evidence‐retrieval F1 or composite relevance metrics.

For the self-consistency mechanism in Stage 1, we leverage stochastic sampling to capture the model’s latent uncertainty. Specifically, we issue R = 5 independent generations per question–note pair, each sampled at temperature 0.7. A majority‐vote over these five runs determines the final label for each sentence, suppressing spurious outliers while preserving genuinely informative evidence. All other decoding parameters (e.g., top-p, frequency and presence penalties) are held at their API defaults, isolating temperature and context length as the principal levers in our experimental configuration.

\section{Prompts and Code Availability}

To promote transparency and reproducibility, we release all manual and optimized prompt templates, together with our full pipeline implementation at our GitHub repository\footnote{\url{https://github.com/ViswanathaReddyGajjala/ArchEHR-QA-Neural}}.

\bibliographystyle{acl_natbib}
\bibliography{anthology,custom}

\end{document}